\documentclass[12pt]{article}
\usepackage{amsfonts}
\usepackage{amsmath}
\usepackage{makeidx}
\usepackage{amssymb}
\usepackage{graphicx}
\usepackage[T1]{fontenc}

\newtheorem{theorem}{Theorem}

\newtheorem{corollary}[theorem]{Corollary}

\newtheorem{adefinition}[theorem]{Definition}

\newtheorem{lemma}[theorem]{Lemma}

\newtheorem{proposition}[theorem]{Proposition}
\newtheorem{aremark}[theorem]{Remark}
\newenvironment{remark}{\begin{aremark}\rm}{\end{aremark}}

\newenvironment{proof}[1][Proof]{\noindent\textbf{#1.} }{\ \rule{0.5em}{0.5em}}
\numberwithin{equation}{section} \numberwithin{theorem}{section}
\begin{document}

\title{Eigenvalue Distribution\\
of Large Random Matrices \\
Arising in Deep Neural Networks: \\
Orthogonal Case}
\author{L. Pastur \\
B.Verkin Institute for Low Temperature Physics and Engineering\\
Kharkiv, Ukraine}

\date{}
\maketitle

\begin{abstract}
The paper deals with the distribution of singular values of the input-output
Jacobian of deep untrained neural networks in the limit of their infinite
width. The Jacobian is the product of random matrices where the independent
rectangular weight matrices alternate with diagonal matrices whose entries
depend on the corresponding column of the nearest neighbor weight matrix. The problem was
considered in \cite{Pe-Co:18} for the Gaussian weights and biases and also for the weights that are Haar distributed orthogonal matrices and Gaussian biases. Basing on a
free probability argument, it was claimed
that in these
cases the singular value distribution of the Jacobian in the limit of infinite width (matrix
size) coincides with that of the analog of the Jacobian with
special random but weight independent diagonal
matrices, the case well known in random matrix theory. The claim was
rigorously proved in \cite{Pa-Sl:21} for a quite general class of weights
and biases with i.i.d. (including Gaussian) entries by using a version of the
techniques of random matrix theory. In this paper we use another version of
the techniques to justify the claim for random Haar distributed weight
matrices and Gaussian biases. 
\end{abstract}


\section{Introduction}

Artificial neural networks is an emerging and quite efficient technique with
a wide variety of applications, see, e.g.
\cite{Ba-Co:20,Bu:17,Ca:20,Go-Co:16,Le-Co:15,Ma-Ma:17,Sc:15,Sh-Ma:19}. One of the  basic
ingredients of the networks  is the iterative scheme

\begin{equation}
x^{l}:=\{x_{j_{l}}^{l}\}_{j_{l}=}^{n_{l}},\;x_{j_{l}}^{l}=\varphi
(y_{j_{l}}^{l}),\;y^{l}:=W^{l}x^{l-1}+b^{l},\;l=1, \dots,L,  \label{rec}
\end{equation}%
where
\begin{equation}
x^{0}:=\{x_{j_{0}}^{0}\}_{j_{0}=1}^{n_{0}}\in \mathbb{R}^{n_{0}}  \label{x0}
\end{equation}%
is the data \textit{input} to the network,
\begin{equation}\label{xl}
x^{L}:=\{x_{j_{L}}^{L}%
\}_{j_{L}=1}^{n_{L}}\in \mathbb{R}^{n_{L}}
\end{equation}
is its \textit{output},
\begin{equation}
W^{l}:=\{W_{j_{l}j_{l-1}}^{l}\}_{j_{l},j_{l-1}=1}^{n_{l},n_{l-1}},\;l=1,
\dots , L  \label{wl}
\end{equation}%
are $n_{l}\times n_{l-1}$ rectangular \textit{synaptic weight} matrices,
\begin{equation}
b^{l}:=\{b_{j_{l}}^{l}\}_{j_{l}=1}^{n_{l}},\;l=1, \dots ,L  \label{bl}
\end{equation}%
are $n_{l}$-component \textit{bias} vectors of the $l$th \textit{layer}, $%
n_{l}$ is the \textit{width} of the $l$th layer\textit{\ }, $\varphi :%
\mathbb{R}\rightarrow \mathbb{R}$ is the component-wise \textit{nonlinearity}
(\textit{activating function}) and $L$ is the \textit{depth} of the network.
If $L>1$, then it is a \emph{deep} neural network (DNN).

Another basic ingredient of the DNN is the \textit{training}
  which modifies the \emph{parameters} (weight matrices $W^l$ and biases $b^l$) on the every step of the iterative scheme
in order to reduce the misfit between the output data and the prescribed
data by using a certain optimization procedure (based often on a version of
least square method). Being multiply repeated in the DNN, the procedure provides
an output $x^L$ (a recognized pattern, a translated text, etc.) but also
 certain final parameters  $W^L$ and $b^L$, which could have a quite non-trivial
structure, see e.g. \cite{Ma-Ma:17}.

It is important for this paper that the theory deals also with untrained and/or random parameters of the DNN architecture, see, e.g. \cite%
{Ca-Sc:20,Gi-Co:16,Li-Qi:00,Ma-Co:16,Pe-Co:18,Po-Co:16,
Sc-Wa:17,Sc-Co:17,Ta-Co:18,Ya:20} and references therein. It is
assumed in these works that the weight matrices and the bias vectors are
independent and identically distributed (i.i.d.) in $l$.

An important characteristic  of the DNN is the input-output Jacobian (see \cite%
{Pe-Co:18} and references therein)
\begin{equation}
J_{\mathbf{n}_L}^{L}:=\left\{ \frac{\partial x_{j_{L}}^{L}}{\partial
x_{j_{0}}^{0}}\right\}
_{j_{0},j_{L}=1}^{n}=\prod_{l=1}^{L}D_{n_{l}}^{l}W_{n_{l}}^{l},\; \; \mathbf{n}_L=\{n\}_{l=1}^{L},  \label{jac}
\end{equation}%
the $n_{L}\times n_{0}$ random matrix, where
\begin{equation}
D_{n_{l}}^{l}:=\{D_{j_{l}}^{l}\delta
_{j_{l}k_{l}}\}_{j_{l},k_{l}=1}^{n_{l}},\;D_{j_{l}}^{l}:=\varphi ^{\prime }%
\Big(\sum_{j_{l-1}=1}^{n}W_{j_{l}j_{l-1}}^{l}x_{j_{l-1}}^{l-1}+b_{j_{l}}^{l}%
\Big),\;l=1,\dots,L  \label{D}
\end{equation}%
are diagonal random matrices and $\mathbf{n_L}:=\{n_{l}\}_{l=1}^{L}$ .

Having in mind that the widths $\mathbf{n_L}:=\{n_{l}\}$ of layers are usually large, one
looks for the characteristics of Jacobian that are well defined in this
asymptotic regime.
Since the spectral properties of 
the Jacobian are strongly correlated with the success of training, one of such
characteristics is the distribution of the singular values of $J_{\mathbf{n}%
_{L}}^{L} $, i.e., the square roots of eigenvalues of the $n_{L}\times n_{L}$
positive definite matrix
\begin{equation}
M_{\mathbf{n}_{L}}^{L}:=J_{\mathbf{n}_{L}}^{L}(J_{\mathbf{n}_{L}}^{L})^{T},
\label{JJM}
\end{equation}%
for networks with random weights and biases and for large widths of layers, see \cite%
{Li-Qi:00,Ma-Co:16,Pe-Ba:17,Pe-Co:18,Po-Co:16,Sc-Co:17,Ta-Co:18,Ya:20} for
various motivations, settings and results. 
More precisely, one studies the asymptotic regime 
determined by the simultaneous limits
\begin{equation}
\lim_{N_{l}\rightarrow \infty }\frac{n_{l-1}}{n_{l}}=c_{l}\in (0,\infty
),\;n_{l}\rightarrow \infty ,\;l=1,\dots ,L.  \label{asf1}
\end{equation}%
%
%
%
Note, however, that many principal results and difficulties in their proofs
are practically the same for the case of distinct $n_{l}\rightarrow \infty
,\;l=1,\dots, L$ in (\ref{asf1}) and for that where
\begin{eqnarray}\label{ns}
n :=n_{l}=\dots =n_{L}.
\end{eqnarray}%
Thus, we confine ourselves to this case writing everywhere below $n$ instead
of $\mathbf{n}_{L}$.


Denote $\{\lambda _{t}^{L}\}_{t=1}^{n}$ the eigenvalues of the random matrix
(\ref{JJM}) and introduce its \emph{ Normalized Counting Measure (NCM)}
\begin{equation}
\nu _{M_{n}^{L}}:=n^{-1}\sum_{t=1}^{n}\delta _{\lambda _{t}^{L}}.  \label{ncm}
\end{equation}%
We will deal with the limit 
\begin{equation}
\nu _{M^{L}}:=\lim_{n\rightarrow \infty }\nu _{M_{n}^{L}}.  \label{ids}
\end{equation}%
Note that since $\nu _{M_{n}^{L}}$ is a random measure, the meaning of the limit has
to be indicated.

\medskip
The problem has been considered in \cite{Pe-Co:18} (see also \cite%
{Ba-Co:20,Hu-Co:20,Li-Qi:00,Pa:20,Pe-Ba:17,Ta-Co:18,Ya:20}) for two cases:

\smallskip (i) $b^{l},$ $l=1,2,\dots ,L$ are $n$-component random vectors i.i.d. in $l$  and having i.i.d. Gaussian components and
$W^{l},\;l=1,2,\dots ,L$ are $n \times n$ random matrices  i.i.d. in $l$  and having
i.i.d. Gaussian entries (see \cite{Pa-Sl:21} for more general i.i.d. components
and entries);

\smallskip (ii) $b^{l},\;l=1,2,\dots ,L$ are as in (i) and
\begin{equation}
W^{l}=O_{n}^{l},\;l=1,2,\dots ,L,  \label{wol}
\end{equation}%
where $O_{n}^{l}\in SO(n),\;l=1,2,\dots ,L$ are $n\times n$ random
orthogonal matrices independent in $l$ and having the normalized to unity
Haar measure on $SO(n)$ as its probability measure.

\medskip This paper deals with the case (ii).

In \cite{Pe-Co:18}
compact formulas for the limit of the expectation
\begin{equation}
\overline{\nu }_{M^{L}}:=\lim_{n\rightarrow \infty }\overline{\nu }%
_{M_{n}^{L}},\;\overline{\nu }_{M_{n}^{L}}:=\mathbf{E}\{\nu _{M_{n}^{L}}\}
\label{mids}
\end{equation}%
of the NCM (\ref{ncm}) and its Stieltjes transform
\begin{equation}
f_{M^{L}}(z):=\int_{0}^{\infty }\frac{\overline{\nu }_{M^{L}}(d\lambda )}{%
\lambda -z},\; z\in\mathbb{C}\setminus \mathbb{R}_+  \label{stm}
\end{equation}%
were proposed for both cases (i) and (ii) above.
The formula for $\overline{\nu }_{M^{L}}$  in the case (ii) is given in (\ref%
{nucon}) below. To write the formula for $f_{M^{L}}$ in the case (ii) it is convenient to
introduce the moment generating function
\begin{equation}
m_{M^{L}}(z):=\sum_{k=1}^{\infty }m_{k}z^{k},\;m_{k}:=\int_{0}^{\infty
}\lambda ^{k}\overline{\nu }_{M^{L}}(d\lambda ),  \label{mgen}
\end{equation}%
of $\overline{\nu }_{M^{L}}$ related to $f_{M^{L}}$ as
\begin{equation}
m_{M^{L}}(z)=-1-z^{-1}f_{M^{L}}(z^{-1}).  \label{stmg}
\end{equation}%
Let
\begin{equation}
K_{n}^{l}:=(D_{n}^{l})^{2}=\{(D_{j_{l}}^{l})^{2}\}_{j_{l}=1}^{n}  \label{kan}
\end{equation}%
be the square of the $n_{l}\times n_{l}$ random diagonal matrix (\ref{D})
and let $m_{K^{l}}$ be the moment generating function of the $n\rightarrow
\infty $ limit $\overline{\nu }_{K^{l}}$ of the expectation $\overline{\nu }_{K_{n}^{l}}$ of the NCM $\nu _{K_n^{l}}$ of $%
K_{n}^{l}$ (see (\ref{nukal})). Then we have according to formulas (14) and (16) in \cite%
{Pe-Co:18} in the case where $\overline{\nu }_{K^{l}}$, hence, $m_{K^{l}}$
do not depend on $l$ (see Remark \ref{r:penn} (i))
\begin{align}
m_{M^{L}}(z)& =m_{K}(z^{1/L}\Psi _{L}(m_{M^{L}}(z))),  \notag \\
\Psi _{L}(z)& =((1+z)/z)^{1-1/L}.  \label{penfo}
\end{align}
In other words, $m_{M^{L}}$, hence, $f_{M^{L}}$ of (\ref{stm}) -- (\ref{stmg}),  
satisfy a certain functional equation, the standard situation in random
matrix theory and its applications, see e.g. \cite{Gi:12,Ma-Pa:67,Pa:72,Pa:00,Pa-Sh:11} and
formulas (\ref{fhk}) -- (\ref{kh}) below.
Note that our notation is different from that of \cite{Pe-Co:18}: our $%
f_{M^{L}}(z)$ of (\ref{stm}) is $-G_{X}(z)$ of (7) in \cite{Pe-Co:18} and
our $m_{M^{L}}(z)$ of (\ref{mgen}) is $M_{X}(1/z)$ of (9) in \cite{Pe-Co:18}.

These and other related results are obtained in \cite{Pe-Co:18}
by using the claimed in this work asymptotic
freeness of the diagonal matrices $D_{n}^{l},\;l=1,\ldots ,L$ of (\ref{D})
and the matrices $O^{l},\;l=1,\ldots ,L$ of (\ref{rec}) and (\ref{wol})
(see, e.g. \cite{Ch-Co:18,Mi-Sp:17} for the definitions and properties of
asymptotic freeness). This leads directly to (\ref{nucon}) and (\ref{penfo})
in view of the multiplicative property of the moment generating functions (%
\ref{mgen}) and the so-called $S$-transforms of $\overline{\nu }_{K^{l}}$
and of $\nu _{O^{l}(O^{l})^{T}}$, the mean limiting NCM's of $K_{n}^{l}$ and
of $O_{n}^{l}(O_n^{l})^{T}$, also see Remark \ref{r:penn} (i) and Corollary \ref%
{c:conv}. 

There is, however, a subtle point in the claim made in \cite{Pe-Co:18}.
Indeed, to the best of our knowledge the asymptotic freeness has been
established so far for the random Gaussian and orthogonal random matrices $%
W^{l}$ and deterministic (more generally, random but $W^{l}$-independent)
diagonal matrices, see e.g. the recent book \cite{Mi-Sp:17}, Chapters 1 and
4. On the other hand, the diagonal matrices $D_{n}^{l}$ in (\ref{D}) depend
explicitly on $(W^{l},b_{n}^{l})$ of (\ref{wl}) -- (\ref{bl}) and,
implicitly, via $x^{l-1}$, on the all "preceding" $(W^{l^{\prime
}},b_{n}^{l^{\prime }}),\;l^{\prime }=l-1,\dots ,1 $. Thus, the proof of
validity of (\ref{nucon}) and (\ref{penfo}) requires an additional
argument. It was given in \cite{Pa:20} for the Gaussian weights and
biases and in \cite{Pa-Sl:21} for a wide class
of  weights and biases with  i.i.d. but not necessarily Gaussian entries
and components. 
In this paper we 
justify the results of \cite{Pe-Co:18} for the orthogonal weights of (%
\ref{wol}) and Gaussian biases with i.i.d. components (for more general biases see Remark \ref%
{r:clt} below).


It is also worth noting that we  prove that formula (\ref{ids}) is valid not
only in the mean (see (\ref{mids}) and \cite{Pe-Co:18}), but also with
probability 1 (recall that the measures in the r.h.s. of (\ref{ids} are
random) and that the limiting measure $\nu _{M^{L}}$ in the l.h.s. of (\ref%
{ids}) coincides with $\overline{\nu }_{M^{L}}$ of (\ref{mids}), i.e., $\nu
_{M^{L}}$ is non-random.

Our approach is an updated version of that developed in \cite{Pa-Sh:11}, Sections
8 --10 for the
spectral analysis of random matrices whose randomness is due the classical
compact groups viewed as probability spaces with the normalized to unity
Haar measure. This shows that the matrices (\ref{JJM}) are in the scope of
random matrix theory, especially  that part of the theory which was created
by Dyson in the 1960s. Note that the justification of results of \cite{Pe-Co:18}
for matrices with Gaussian and more general i.i.d. weight entries and bias
components, given in \cite{Pa:20,Pa-Sl:21}, is bazed an appropriately updated version of the tools of
random matrix
theory presented in \cite{Pa-Sh:11}, Chapter 7, 18 and 19, although this
version
and that of this paper are borrowed from different parts of random matrix theory and
have not too much in common.

An additional motivation of the paper is that according to \cite{Pe-Co:18}
the tight concentration of the entire spectrum of singular values of the
input-output Jacobian (\ref{jac}) around the point 1 of the spectral axis
can considerably enhance the efficiency of the network in question, especially
on the initial steps of iteration procedure (see also \cite{Hu-Co:20,Li-Qi:00,Ta-Co:18}), and that
the orthogonal weights provide this property most simply.

The paper is organized as follows. In the next section we prove the validity
of (\ref{ids}) with probability 1, formulas (\ref{nucon}) for $\nu _{M^{L}}=%
\overline{\nu }_{M^{L}}$ and (\ref{penfo}) of \cite{Pe-Co:18}. The proof is
based on a natural inductive procedure allowing for the passage from the $l$%
th to the $(l+1)$th layer and it is fairly similar to that in \cite%
{Pa:20,Pa-Sl:21}. This is because the passage procedure is almost
independent of the probability properties of the weight entries provided
that a formula relating the limiting (in the layer width) Stieltjes
transforms of the NCM's of two subsequent layers is known. This formula and
a number of auxiliary results are proved in Section 3.

Note that to make the paper self-consistent we present here certain facts
and arguments that have been already given in \cite{Pa:20,Pa-Sl:21}, thus
the paper is in part a review of these works. 


\section{Main Result and its Proof.}

As was already mentioned in Introduction, the goal of the paper is to
justify the results of \cite{Pe-Co:18} for the independent in $l$ and the
Haar distributed orthogonal weights $O^{l}$'s and the independent in $l$
biases $b^{l}$'s with independent  Gaussian components (see Remark \ref%
{r:clt} for more general biases).

More precisely, we consider the case of (\ref{x0}) -- (\ref{bl}) where:

\smallskip (i) all $b_{n}^{l}$ and $W_{n}^{l},\;l=1,\dots ,L$ in (\ref{bga})
-- (\ref{wga}) are of the same size $n$ and $n\times n$ respectively, i.e., (%
\ref{ns}) holds true; 

\medskip (ii) the bias vectors $b_{n}^{l}=\{b_{j_{l}}^{l}\}_{j_{l}=1}^{n},%
\;l=1,2,\dots ,L$ are random i.i.d. in $l$ and for every $l$ their
components are i.i.d. Gaussian random variables with
\begin{equation}
\mathbf{E}_{b^{l}}\{b_{j_{l}}^{l}\}=0,\;\mathbf{E}_{b^{l}}%
\{(b_{j_{l}}^{l})^{2}\}=\sigma _{b}^{2}>0,  \label{bga}
\end{equation}%
where $\mathbf{E}_{b^{l}}\{\ldots \}$ denotes the expectation in the
probability space of $b^{l}$;

(iii) the weight matrices $W_{n}^{l},\;l=1,2,\ldots ,L$ are also i.i.d in $l$
and for every $l$
\begin{equation}
W_{n}^{l}=O_{n}^{l}=\{O_{j_{l}j_{l-1}}^{l}\}_{j_{l}j_{l-1}=1}^{n},
\label{wga}
\end{equation}%
where $O_{n}^{l}$ is the random matrix with values in the group $SO(n)$ of orthogonal
and unimodular ($\det O_{n}^{l}=1$) matrices. The group plays the role  of
the probability space and the
normalized to unity Haar measure on the group plays the role of the probability
measure. In
particular, we have%
\begin{equation}
\mathbf{E}_{O^{l}}\{O_{j_{l}j_{l-1}}^{l}\}=0,\;\mathbf{E}_{O^{l}}%
\{O_{j_{l_{1}}j_{l_{1}-1}}^{l}O_{j_{l_{2}}j_{l_{2}-1}}^{l}\}=n^{-1}%
\delta _{j_{l_{1}}j_{l_{2}}}\delta
_{j_{l_{1}}-1,\, j_{l_{2}-1}},  \label{wga1}
\end{equation}%
where $\mathbf{E}_{O^{l}}\{\dots \}$ denotes the expectation with respect to
the normalized to unity Haar measure on $SO(n)$. 

For every $l$ we view $b_{n}^{l}$ as the first $n$ components of the
semi-infinite random vector
\begin{equation}
\{b_{j_{l}}^{l}\}_{j_{l}=1}^{\infty }  \label{binf}
\end{equation}%
whose independent Gaussian components satisfy (\ref{bga}) and we denote $%
\Omega _{b_{l}}$ the infinite-dimensional (product) probability space for (%
\ref{binf}).

Next, it follows from Proposition \ref{p:facts} (iii)
that there exists an analogous infinite-dimensional space for the sequence
\begin{equation}
\{O_{n}^{l}\}_{n=1}^{\infty }.  \label{xinf}
\end{equation}%
We denote this space by $\Omega _{O_{l}}$ and by $\mathbf{E}_{O_{l}}\{\ldots
\}$ the expectation in this space.

As a result of the above construction of the infinite-dimensional
probability spaces for weights and biases of the $l$th layer they are now
defined for all $n=1,2,\dots $ \ on the same infinite-dimensional product
probability space
\begin{equation}\label{oob}
\Omega ^{l}=\Omega _{b_{l}}\times \Omega _{O_{l}}.
\end{equation}
Let
also
\begin{equation}\label{oml}
\Omega _{L}=\Omega ^{L}\times \Omega ^{L-1}\times \dots \times \Omega
^{1}
\end{equation}%
be the infinite-dimensional probability space on which the recurrence (\ref%
{rec}) is defined for a given depth $L$. This will allow us to formulate
various results on the large size asymptotic behavior of the eigenvalue
distribution of matrices (\ref{JJM}) as those valid with probability 1 in $%
\Omega _{L}$. We will denote $\mathbf{E}\{\ldots \}$ the expectation in $%
\Omega _{L}$.

In fact, it was argued in \cite{Pe-Co:18} for the Gaussian and the
orthogonal weights (and proved in \cite{Pa:20,Pa-Sl:21} for the weights
with the Gaussian and the i.i.d. entries) that the resulting eigenvalue
distribution of random matrices (\ref{JJM}) coincides with that of matrices
of the same form where, however, the analogs of diagonal matrices (\ref{D})
are random \textit{but independent} of $W^{l}$. In this paper we prove an
analogous result for orthogonal weighs. Thus, we formulate first the
corresponding results of random matrix theory which are largely known (see,
e.g. \cite{Pa-Sh:11}, Section 10.4, \cite{Va:01}, \cite{Pa-Va:22} and references therein)

Consider for every positive integer $n$: (i) the $n\times n$ random Haar
distributed over the group matrices $O_{n}\in O(n)$ (see (\ref{wga}) -- (\ref{wga1}%
)) and defined for all $n$ of the same probability space $\Omega_O$
(see (\ref{xinf})); (ii) the $n\times n$ positive definite matrices $\mathsf{K}_{n}$ and $%
\mathsf{R}_{n}$ (that may also be random but independent of $O_{n}$ and defined
on the same probability space $\Omega_{KR}$ for all $n$ (cf. (\ref{xinf}))) and such
that their Normalized Counting Measures $\nu _{\mathsf{K}_{n}}$ and $\nu _{%
\mathsf{R}_{n}}$ (see (\ref{ncm})) converge weakly (with probability 1 if
random) as $n\rightarrow \infty $ to non-random measures $\nu _{\mathsf{K}}$
and $\nu _{\mathsf{R}}$:%
\begin{equation}
\nu _{\mathsf{K}_{n}}\rightarrow \nu _{\mathsf{K}},\;\nu _{\mathsf{R}%
_{n}}\rightarrow \nu _{\mathsf{R}},\;n\rightarrow \infty .  \label{lnkr}
\end{equation}%
Set
\begin{equation}
\mathsf{M}_{n}=\mathsf{K}_{n}^{1/2}O_{n}\mathsf{R}_{n}O_{n}^{T}\mathsf{K}%
_{n}^{1/2}.  \label{mnsf}
\end{equation}%
According to random matrix theory
(see, e.g. \cite{Pa-Sh:11}, Section 10.4,
 \cite{Pa-Va:22,Va:01} and Lemma \ref{l:vas} below), in this
case and under certain conditions on $\mathsf{K}_{n}$ and $\mathsf{R}_{n}$
the Normalized Counting Measure $\nu _{\mathsf{M}_{n}}$ of $\mathsf{M}_{n}$
converges weakly with probability 1 (on $\Omega_O \times \Omega_{KR}$) as $n\rightarrow \infty $ to a
non-random measure $\nu _{\mathsf{M}}$ which is uniquely determined by the
limiting measures $\nu _{\mathsf{K}}$ and $\nu _{\mathsf{R}}$ of (\ref{lnkr}%
) via a certain analytical procedure (see, e.g. formulas (\ref{stm}) and (%
\ref{fhk}) -- (\ref{kh}) below).

We can write down symbolically this fact as%
\begin{equation}
\nu _{\mathsf{M}}=\nu _{\mathsf{K}}\boxtimes \nu _{\mathsf{R}}  \label{opdia}
\end{equation}%
to stress that the procedure defines a binary operation in the set of
non-negative measures of total mass 1 and of support belonging to the
positive semi-axis (see more in Corollary \ref{c:conv}).
The operation was studied in detail in free probability, \cite{Mi-Sp:17},
having the above random matrices as a basic analytic model, and is know
there as the \emph{free multiplicative convolution}.

It follows from \cite{Pe-Co:18} that the limiting Normalized Counting Measure (\ref%
{ids}) of random matrices (\ref{JJM}), where the role of $\mathsf{K}_{n}$
plays the matrix (\ref{kan}) that depends on matrices $O^{l}$'s of (\ref{wga}%
), can be found as the "product" with respect the operation (\ref{opdia}) of
$L$ measures $\nu _{K^{l}},\;l=1,\dots ,L$ which are the limiting Normalized
Counting Measures of random matrices of (\ref{kan}) - (\ref{D}) given in (%
\ref{nukal}) -- (\ref{qlql}).
This claim can be reformulated as follows. Write (\ref{JJM}) with $%
W^{l}=O_{n}^{l},\,l=1,\ldots ,L$ in (\ref{jac}) -- (\ref{D}) as
\begin{equation}  \label{mlml1}
M_{n}^{l}=D_{n}^{l}O_{n}^{l}M_{n}^{l-1}(O_{n}^{l})^{T}D_{n}^{l}
\end{equation}
and observe that $M_{n}^{l-1}$ is random but \emph{independent} of $%
O_{n}^{l} $, hence can play the role of $R_{n}$ in (\ref{mnsf}). Thus, to be
able to write (\ref{opdia}), we have to assume that $D_{n}^{l}$ of (\ref{D})
can be replaced by%
\begin{equation}
\mathsf{D}_{n}^{l}=\{\mathsf{D}_{j}^{l}\delta _{jk}\}_{j,k=1}^{n},\;\mathsf{D%
}_{j}^{l}=\varphi ^{\prime }\Big((\mathsf{O}_n^{l} \
x^{l-1})_{j}+b_{j_{l}}^{l}\Big),\;l=1,\dots ,L,  \label{dscn}
\end{equation}%
where $\mathsf{O}^{l}_{n},\;l=1,\dots ,L$ are Haar distributed orthogonal
matrices that are \emph{independent} of $O^{l}_{n},\;l=1,\dots ,L$.

The goal of the paper is to justify this replacement in the limit $%
n\rightarrow \infty $ for the widths of layers.

\begin{theorem}
\label{t:main} Let $M_{n}^{L}$ be the random matrix (\ref{JJM}) defined by (%
\ref{rec}) -- (\ref{D}) and (\ref{ns}), where the weights $%
\{O_{n}^{l}\}_{l=1}^{\infty }$ are i.i.d. in $l$ and are $n\times n$ Haar
distributed orthogonal matrices for every $n$
(see (\ref{wga}) -- (\ref{wga1}) and (\ref{xinf})), the biases $%
\{b_{n}^{l}\}_{l=1}^{\infty }$ are i.i.d. in $l$ and are $n$-component
vectors with independent Gaussian components (see (\ref{bga}) and (\ref{binf})) for
every $n$ 
and the input vector $x^{0}$ (\ref{x0}) (deterministic or random) is such
that there exists a finite limit%
\begin{equation}
q^{0}:=\lim_{n\rightarrow \infty }q_{n}^{0}>\sigma
_{b}^{2}>0,\;\;q_{n}^{0}=n^{-1}\sum_{j_{0}=1}^{n}(x_{j_{0}}^{0})^{2}+\sigma
_{b}^{2}.  \label{q0}
\end{equation}%
%
%
%
%
%
%
%
%
%
%
%
%
%
%
%
%
%
%
%
%
Assume also that the $n$-independent nonlinearity $\varphi $ in (\ref{rec}) has a piece-wise
differential derivative $\varphi ^{\prime }$ which is not zero identically
and%
\begin{equation}
\sup_{t\in \mathbb{R}}|\varphi (t)|=\Phi _{0}<\infty ,\;\sup_{t\in \mathbb{R}%
}|\varphi ^{\prime }(t)|=\Phi _{1}<\infty .\;\;  \label{phi1}
\end{equation}%
Then the Normalized Counting Measure (NCM) $\nu _{M_{n}^{L}}$ of $M_{n}^{L}$
(see (\ref{ncm})) converges weakly with probability 1 in the probability
space $\Omega _{L}$ of (\ref{oml}) to the non-random measure
\begin{equation}
\nu _{M^{L}}=\nu _{K^{L}}\boxtimes \cdots \boxtimes \nu _{K^{1}}\boxtimes
\delta _{1},  \label{nucon}
\end{equation}%
where the operation "$\,\boxtimes$" is the free multiplicative convolution,
given in (\ref{opdia}) (see also \cite{Ch-Co:18,Mi-Sp:17} and
Corollary \ref{c:conv}) below), $\delta _{1}$ is the unit measure
concentrated at 1 and
\begin{equation}
\nu _{K^{l}}(\Delta )=\mathbf{P}\{(\varphi ^{\prime }((q^{l-1})^{1/2}\gamma
)^{2}\in \Delta \},\;\Delta \in \mathbb{R},\;l=1,\ldots ,L,  \label{nukal}
\end{equation}%
with the standard Gaussian random variable $\gamma $ and $q^{l}$ determined
by the recurrence
\begin{equation}
q^{l}=(2\pi )^{-1/2}\int_{-\infty }^{\infty }\varphi ^{2}\big(\gamma \sqrt{%
q^{l-1}}\big)e^{-\frac{\gamma ^{2}}{2}}d\gamma ,\;l\geq 1,  \label{qlql}
\end{equation}%
where $q^{0}$ is given by (\ref{q0}).
\end{theorem}

\begin{remark}
\label{r:penn} (i) If
\begin{equation}
q_{L-1}=\cdots =q_{0},  \label{qeq}
\end{equation}%
then $\nu _{K}:=\nu _{K^{l}},\;l=1,\dots ,L$ and (\ref{nucon}) becomes
\begin{equation}
\nu _{M^{L}}=\underset{L\;\mathrm{times}}{\underbrace{\nu _{K}\boxtimes
\cdots \boxtimes \nu _{K}}\diamond \delta _{1}}.  \label{nucone}
\end{equation}
An important case of (\ref{qeq}) is where $q^{0}=q^{\ast }$ and $q^{\ast }$
is a fixed point of (\ref{qlql}), see \cite{Ma-Co:16,Po-Co:16,Sc-Co:17} for
a detailed analysis of (\ref{qlql}) and its role in the deep neural networks
setting.

(ii) If the input vectors (\ref{x0}) are random, then it is necessary to
assume that they are defined on the same probability space $\Omega _{x^{0}}$
for all $n$ and that (\ref{q0}) is valid with probability 1 in $\Omega
_{x^{0}}$, i.e., there exists
\begin{equation}
\overline{\Omega }_{x^{0}}\subset \Omega _{x^{0}},\;\mathbf{P}(\overline{%
\Omega }_{x^{0}})=1  \label{px0}
\end{equation}%
where (\ref{q0}) holds. It follows then from the Fubini theorem that in this
case the set $\overline{\Omega }_{L}\subset \Omega _{L},\,\mathbf{P}\{%
\overline{\Omega }_{L}\}=1$ where Theorem \ref{t:main} holds has to be
replaced by the set $\overline{\Omega }_{Lx^{0}}\subset \Omega _{L}\times
\Omega _{x^{0}},\;\mathbf{P}\{\overline{\Omega }_{Lx^{0}}\}=1$. An example
of this situation is where $\{x_{j}^{0}\}_{j=1}^{n}$ are the first $n$
components of an ergodic sequence $\{x_{j}^{0}\}_{j=1}^{\infty }$ (e.g. a
sequence of i.i.d. random variables) with finite fourth moment. Here $q_{1}$
in (\ref{q0}) exists with probability 1 on the corresponding $\Omega
_{x^{0}} $ and even is non-random just by ergodic theorem (the strong Law of
Large Numbers in the case of i.i.d. sequence) and the theorem is valid with
probability 1 in $\Omega _{L}\times \Omega _{x^{0}}$.%
\end{remark}

We present now the proof of Theorem \ref{t:main}. \medskip

\begin{proof}
We prove the theorem by induction in $L$. We have from (\ref{rec}) -- (\ref%
{JJM}) and (\ref{ns}) with $L=1$ the following $n\times n$ random matrix%
\begin{equation}
M_{n}^{1}=J_{n}^{1}(J_{n}^{1})^{T}=D_{n}^{1}O_{n}^{1}(O_{n}^{1})^{T}D_{n}^{1}=(D_{n}^{1})^{2}:=K_{n}^{1}.
\label{m1}
\end{equation}%
The matrix is a particular case with $R_{n}=\mathbf{1}_{n}$ of matrix (\ref%
{mncal}) treated in Theorem \ref{t:ind} below. Since the NCM of $\mathbf{1}%
_{n}$ is the Dirac measure $\delta _{1}$, conditions (\ref{r2}) -- (\ref{nur}%
) of the theorem are evident. Condition (\ref{qqn}) is just (\ref{q0}).
It follows then from Corollary \ref{c:conv} that the assertion of our
theorem, i.e., formula (\ref{nucon}) with $q^{0}$ of (\ref{q0}), is valid
for $L=1$.

Consider now the case $L=2$ of (\ref{rec}) -- (\ref{JJM}) and (\ref{ns}):
\begin{equation}
M_{n}^{2}=D_{n}^{2}O_{n}^{2}M_{n}^{1}(O_{n}^{2})^{T}D_{n}^{2}.  \label{m2m1}
\end{equation}%
Observe that the matrix is a particular case of matrix (\ref{mncal}) of
Theorem \ref{t:ind} with $M_{n}^{1}$ of (\ref{m1}) as $R_{n}$, $O_{n}^{2}$
as $O_{n}$, $D_{n}^{2}$ as $D_{n}$, $\{x_{j_{1}}^{1}\}_{j_{1}=1}^{n}$ as $%
\{x_{\alpha n}\}_{\alpha =1}^{n}$, $\Omega _{1}=\Omega ^{1}$ of (\ref{oml})
as $\Omega _{Rx}$ and $\Omega ^{2}$ of (\ref{oml}) as $\Omega _{Ob}$, i.e.,
the case of the random but\ $\{O_{n}^{2},b_{n}^{2}\}$ - independent $R_{n}$
and $\{x_{\alpha n}\}_{\alpha =1}^{n}$ in (\ref{mncal}) as described in
Remark \ref{r:rra} (i). Let us check that conditions (\ref{r2}) -- (\ref{nur}%
) \ and (\ref{qqn}) of Theorem \ref{t:ind} are satisfied for $M_{n}^{2}$ of (%
\ref{m2m1}) with probability 1 in the probability space $\Omega _{1}=\Omega
^{1}$ generated by $\{O_{n}^{1},b_{n}^{1}\}$ for all $n$ and independent of
the space $\Omega ^{2}$ generated by $\{O_{n}^{2}, b_{n}^{2}\}$ for all $n$
(see (\ref{oml})).

We will use the bounds:
\begin{equation}
||D_{n}^{1}||\leq \Phi _{1},  \label{nok}
\end{equation}%
following from (\ref{D}) and (\ref{phi1}) and valid everywhere in $\Omega
_{1}$ of (\ref{oml}) and
\begin{equation}
|\mathrm{Tr}AB|\leq ||A||\mathrm{Tr}B,\;  \label{tab}
\end{equation}%
valid for any matrix $A$ and a positive definite matrix $B$. According to
the bounds, we have
everywhere in $\Omega _{1}$:
\begin{equation*}
n^{-1}\mathrm{Tr}(M_{n}^{1})^{2}=n^{-1}\mathrm{Tr}\mathbb{(}%
K_{n}^{1})^{2}\leq \Phi _{1}^{4}.
\end{equation*}%
We conclude that $M_{n}^{1}$, that plays here the role of $R_{n}$ of
Theorem \ref{t:ind} and Remark \ref{r:rra} (i) according to (\ref{m2m1}),
satisfies condition (\ref{r2}) with $r_{2}=\Phi _{1}^{4}$ and with
probability 1 in our case, i.e., on a certain $\Omega _{11}\subset \Omega
_{1},\;\mathbf{P}(\Omega _{11})=1$.

Next, it follows from the above proof of the theorem for $L=1$, i.e., in
fact, from Theorem \ref{t:ind}, that there exists $\Omega _{12}\subset
\Omega _{1},\;\mathbf{P}(\Omega _{12})=1$ on which the NCM $\nu _{M_{n}^{1}}$
converges weakly to a non-random limit $\nu _{M^{1}}$, hence condition (\ref%
{nur}) is also satisfied with probability 1, i.e., on $\Omega _{12}$.

At last, according to Lemma \ref{l:xlyl} with $l=1$ and (\ref{q0}), there
exists $\Omega _{13}\subset \Omega _{1},\;\mathbf{P}(\Omega _{13})=1$ on
which there exists
\begin{equation*}
\lim_{n\rightarrow \infty
}n^{-1}\sum_{j_{1}=1}^{n}(x_{j_{1}}^{1})^{2}+\sigma _{b}^{2}=q^{1}>\sigma
_{b}^{2},
\end{equation*}%
and according to (\ref{rec}) and (\ref{phi1}) we have uniformly in $n$: $%
|x_{j_{1}}^{1}|\leq \Phi _{0},\;j_{1}=1,\dots ,n$, i.e., condition (\ref{qqn}%
) is also satisfied.

Hence, we can apply Theorem \ref{t:ind} and Corollary \ref{c:conv} on the
subspace
$$
\overline{\Omega }_{1}=\Omega _{11}\cap \Omega _{12}\cap \Omega
_{13}\subset \Omega _{1},\;\mathbf{P}
(\overline{\Omega }_{1})=1
$$ where all
the conditions of the theorem are valid, i.e., $\overline{\Omega }_{1}$
plays the role of $\Omega _{Rx}$ of Remark \ref{r:rra} (i). The
theorem implies that for every $\omega _{1}\in \overline{\Omega }_{1}$ there
exists subspace $\overline{\Omega ^{2}}(\omega _{1})$ of the space $\Omega
^{2}$ generated by $\{O_{n}^{2},b_{n}^{2}\}$ for all $n$ and such that $%
\mathbf{P}(\overline{\Omega ^{2}}(\omega _{1}))=1$ and formulas (\ref{nucon}%
) -- (\ref{qlql}) are valid for $L=2$. Then the Fubini
theorem implies that the same is true on a certain $\overline{\Omega }_{2}\subset
\Omega _{2},\;\mathbf{P}(\overline{\Omega }_{2})=1$ where $\Omega _{2}$ is
defined by (\ref{oml}) with $L=2$.

This proves the theorem for $L=2$. The proof for $L=3,4,\dots $ is analogous,
since in general $M_{n}^{L}$ and $M_{n}^{L-1}$ are related by (\ref{mlml1}),
hence $M_{n}^{L-1}$ plays the role of $R_{n}$ of Theorem \ref{t:ind} .
In particular, we have from (\ref{nok}) -- (\ref{tab}) \ with probability 1
on $\Omega _{L}$ of (\ref{oml})
\begin{equation*}
n^{-1}\mathrm{Tr}(M^{L})^{2}\leq \Phi _{1}^{4} \ n^{-1}\mathrm{Tr}%
(M^{L-1})^{2}\leq \Phi_1^{4L} ,\; \; L\geq 1.
\end{equation*}%
If $x_{0}$ is random, then it is necessary to add the argument given in
Remark \ref{r:penn} (ii).

It follows then form Theorem \ref{t:ind} and Corollary \ref{c:conv} that the
binary operation relating the limiting measures $\nu _{M^{L}},\;\nu _{K^{L}}$
and $\nu _{M^{L-1}}$, hence, implying (\ref{nucon}), is indeed the free
multiplicative convolution given in (\ref{mnsf}) -- (\ref{opdia}).

The derivation of the functional equation (\ref{penfo}) is given in \cite%
{Pe-Co:18}.
\end{proof}

\medskip  Note that the above part of this section is rather close
to that of Section 2 of \cite{Pa-Sl:21} and is given here to make the paper
more self-consistent.

An important property of the network Jacobian (\ref{jac}) is the tight
concentration of its singular value spectrum, i.e., the spectrum of (\ref%
{JJM}), around the point 1 of the spectral axis. This property of DNN is known as
the \emph{dynamical isometry}  and implies that the corresponding
Jacobian is well-conditioned, see \cite{Hu-Co:20,Li-Qi:00,Pe-Co:18,Ta-Co:18}
and references therein. It is indicated in these works that the networks
with the Gaussian weights $W^{l}$'s do not possess this property while the
networks with orthogonal weights and certain non-linearities can achieve the
dynamical symmetry as the depth $L$ increases.

\section{Auxiliary Results.}

Theorem \ref{t:main} of previous section is proved by induction in the depth
$L$ of the network, see formulas (\ref{m2m1}) and (\ref{mlml1}). To pass
from the depth $(L-1)$ to that $L$ we need a formula relating the limiting
NCM $\nu _{M^{L}}$ of the matrix $M_{n}^{L}$ and that $\nu _{M^{L-1}}$ of $%
M_{n}^{L-1}$ in the infinite width limit $n\rightarrow \infty $. The
corresponding result, Theorem \ref{t:ind}, which could be of independent
interest, as well as certain auxiliary results are proved in this section.
In particular, functional equations relating the Stieltjes transforms of $\nu
_{M_{n}^{L}}$ and $\nu _{M_{n}^{L-1}}$ in the limit $n\rightarrow \infty $
are obtained (see (\ref{fhk}) -- (\ref{kh})).

\begin{theorem}
\label{t:ind} Consider for every positive integer $n$ the $n\times n$ random
matrix
\begin{equation}
M_{n}=D_{n}O_{n}R_{n}O_{n}^{T}D_{n},  \label{mncal}
\end{equation}%
where:

\smallskip (a) $R_{n}$ is a positive definite $n\times n$ matrix such that
\begin{equation}
\sup_{n}n^{-1}\mathrm{Tr}R_{n}^{2}=r_{2}<\infty  \label{r2}
\end{equation}%
and
\begin{equation}
\lim_{n\rightarrow \infty }\nu _{R_{n}}=\nu _{R},\;\nu _{R}(\mathbb{R}%
_{+})=1,  \label{nur}
\end{equation}%
where $\nu _{R_{n}}$ is the Normalized Counting Measure of $R_{n}$, $\nu
_{R} $ is a non-negative measure not concentrated at zero and $%
\lim_{n\rightarrow \infty }$ denotes the weak convergence of probability
measures;

\smallskip (b) $O_{n}$ is the $n\times n$ orthogonal Haar distributed random
matrix (see (\ref{wga})), $b_{n}$ is the $n$-component random vector
\begin{equation}
b_{n}=\{b_{j}\}_{j=1}^{n},\;\mathbf{E}\{b_{j}\}=0,\;\mathbf{E}\{b_{j}\}=\sigma
_{b}^{2}  \label{b}
\end{equation}%
with independent Gaussian components (see (\ref{bga})) 
and for all $n$ the matrices $O_{n}$ and the vectors $b_{n}$ are viewed as
defined on the same probability space%
\begin{equation}
\Omega _{Ob}=\Omega _{O}\times \Omega _{b},  \label{oob1}
\end{equation}%
where $\Omega _{O}$ and $\Omega _{b}$ are generated by (\ref{xinf}) and (\ref{binf});

\smallskip (c) $D_{n}$ is the diagonal random matrix
\begin{equation}
D_{n}=\{\delta _{jk}D_{jn}\}_{j,k=1}^{n},\;D_{jn}=\psi \Big(%
(O_{n}x_{n})_{j}+b_{j}\Big),  \label{Dn}
\end{equation}%
where $\psi :\mathbb{R}\rightarrow \mathbb{R}$ is a piecewise
continuous  function that is not identically zero and such that
\begin{equation}
\sup_{x\in \mathbb{R}%
}|\psi (x)|=\Psi<\infty ,  \label{Phi}
\end{equation}%
and the collection%
\begin{equation}
x_{n}=\{x_{\alpha n}\}_{\alpha =1}^{n}\in \mathbb{R}_{n}  \label{xn}
\end{equation}%
admits the limit%
\begin{equation}
q=\lim_{n\rightarrow \infty }q_{n}>\sigma
_{b}^{2}>0,\;q_{n}=n^{-1}\sum_{\alpha =1}^{n}(x_{\alpha n})^{2}+\sigma
_{b}^{2}.  \label{qqn}
\end{equation}%
Denote (cf. (\ref{kan})%
\begin{equation}
K_{n}=D_{n}^{2}.  \label{kndn}
\end{equation}

Then the Normalized Counting Measure (NCM) $\nu _{M_{n}}$ of $M_{n}$
converges weakly with probability 1 in $\Omega _{Ob}$ of (\ref{oob1}) to a
non-random measure $\nu _{M}$, such that $\nu _{M}(\mathbb{R})=1$ and its
Stieltjes transform
\begin{equation}\label{fm}
f_{M}(z)=\int_{0}^{\infty }\frac{\nu _{M}(d\lambda )}{\lambda -z},\;z\in
\mathbb{C}\setminus \mathbb{R}_{+}
\end{equation}%
can be obtained from a unique solution $(f_{M},h_{K},h_{R})$ of the system
of functional equations%
\begin{align}
(1+zf_{M}(z))f_{M}(z)-h_{K}(z)h_{R}(z)& =0,  \label{fhk} \\
f_{R}(zf_{M}(z)/h_{K}(z))& =h_{K}(z),  \label{hk} \\
f_{K}(zf_{M}(z)/h_{R}(z))& =h_{R}(z),  \label{kh}
\end{align}%
where $f_{R}$ and $f_{K}$ are the Stieltjes transforms of $\nu _{R}$ and $%
\nu _{K}$, $\nu _{R}$ is defined in (\ref{nur}) and 
\begin{equation}
\nu _{K}(\Delta )=\mathbf{P}\{(\psi(q^{1/2}\gamma
+b_{1}))^{2}\in \Delta \},\;\Delta \in \mathbb{R},  \label{nuka}
\end{equation}%
with $q$ is given by (\ref{qqn}), $\gamma $ is the standard Gaussian random
variable. The system (\ref{fhk}) -- (\ref{kh}) is uniquely solvable in the
class of triple ($f_{M},h_{K},h_{R}$) of functions analytic outside the
closed positive semi-axis, continuous and positive on the negative semi-axis
and such that%
\begin{equation}
\Im f(z)\Im z>0,\;\Im z\neq 0;\;\sup_{\xi \geq 1}\xi f(-\xi )\in (0,\infty
),\;f=f_{M},h_{K},h_{R}.  \label{hcond}
\end{equation}
\end{theorem}


\begin{remark}
\label{r:rra} To apply Theorem \ref{t:ind} to the proof of Theorem \ref%
{t:main} we need a version of the former in which its "parameters",
i.e., $R_{n}$ in (\ref{mncal}) -- (\ref{nur}), (possibly) $\{x_{\alpha
n}\}_{\alpha =1}^{n}$ in (\ref{Dn}) and $q$ in (\ref{qqn}) are random,
defined for all $n$ on the same probability space $\Omega _{Rx}$,
independent of $\Omega _{Ob}$ of (\ref{oob1}) and satisfying  conditions (\ref{r2}%
) -- (\ref{nur}) and (\ref{qqn}) with probability 1 in $\Omega _{Rx}$, i.e.,
on a certain subspace (cf. (\ref{px0}))
\begin{equation}
\overline{\Omega }_{Rx}\subset \Omega _{Rx},\;\mathbf{P}(\overline{\Omega
_{Rx}})=1.  \label{px}
\end{equation}%
In this case Theorem \ref{t:ind} is valid with probability 1 in $\Omega
_{Ob}\times \Omega _{Rx}$. The corresponding argument is standard in random
matrix theory, see, e.g. of \cite{Pa-Sh:11}), Section 2.3 and Remark \ref%
{r:penn} (ii). 
The obtained limiting NCM\ $\nu _{M}$ is random in general due to the
(possible) randomness of $\nu _{R}$ and $q$ in (\ref{nur}) and (\ref{qqn})
which are defined on the probability space $\Omega _{Rx}$ (but do not depend
on $\omega \in \Omega _{Ob}$). 
Note, however, that in the case of Theorem \ref{t:main} the analogs of $\nu
_{R}$ and $q$ are not random, thus the limiting measure $\nu _{M^{L}}$ is
non-random as well.

\end{remark}

\begin{proof}
Write (\ref{mncal}) as%
\begin{equation}
M(O_{n}):=M(O_{n},D_{n}(O_{n}x_{n}+b_{n}),R_{n})  \label{mo}
\end{equation}%
and replace $O_{n}$ by $O_{n}O_{n}(x)$, where%
\begin{equation}
O_{n}(x_{n})x_{n}=||x_{n}||e_{n},\;x_{n}=||x_{n}||O_{n}^{T}(x_{n})e_{n},%
\;||x_{n}||^{2}=\sum_{\alpha =1}^{n}(x_{j\alpha })^{2},  \label{oxe}
\end{equation}%
with $e_{n\text{ }}$being the $n$th (last) vector of the canonical basis of $%
\mathbb{R}^{n}$.

Because of the orthogonal invariance of the Haar measure on $SO(n)$ the
probability laws of $M_{n}(O_{n})$ and $M_{n}(O_{n}O_{n}(x_{n}))$ coincide
for any $x_{n}\in \mathbb{R}^{n}$. Thus, these matrices are statistically
equivalent, i.e, \ all their statistical characteristics (various moments,
the convergence with probability 1, etc.) are the same for any $x_{n}$. We \
will write this fact as%
\begin{equation}
M_{n}(O_{n})\circeq M_{n}^{(1)}:=M_{n}(O_{n}O_{n}(x_{n})).  \label{steq}
\end{equation}%
In particular, this is the case for the NCM's and of $M_{n}(O_{n})$ and $%
M_{n}^{(1)}$%
\begin{equation}
\nu _{M_{n}(O)}\circeq \nu _{M_{n}^{(1)}},  \label{nunu1}
\end{equation}%
hence, it suffices to prove the convergence of $\nu _{M_{n}^{(1)}}$ with
probability 1.

We have then from (\ref{mncal}) and (\ref{steq}):%
\begin{align}
M_{n}^{(1)}& =M_{n}(O_{n},D_{n}(||x_{n}||O_{n}e_{n}+b_{n}),R_{n}^{O_{n}(x)})
\notag \\
R_{n}^{O_{n}(x_{n})}& =O_{n}(x)R_{n}O_{n}^{T}(x_{n}).  \label{mo1}
\end{align}%
We use now Proposition \ref{p:facts} (i) -- (ii) to present $O_{n}$ as the
product $V_{n}\mathcal{O}_{n}$ (\ref{orep}) of two independent orthogonal matrices $%
\mathcal{O}_{n}$ and $V_{n}$ allowing us to write (\ref{mo1}) as
\begin{align}
M_{n}^{(1)}& =V_{n}M_{n}^{(2)}V_{n}^{T},  \notag \\
M_{n}^{(2)}& =M_{n}(\mathcal{O}%
_{n},D_{n}^{V}(||x_{n}||V_{n}e_{n}+b_{n}),R_{n}^{O_{n}(x_{n})}),\;  \notag \\
D_{n}^{V_{n}}& =V_{n}^{T}D_{n}(||x_{n}||V_{n}e_{n}+b_{n})V_{n}.  \label{mo2}
\end{align}%
Since $M_{n}^{(1)}$ and $M_{n}^{(2)}$\ are orthogonally equivalent, their
spectra, hence, their NCM's, coincide:%
\begin{equation}
\nu _{M_{n}^{(1)}}=\nu _{M_{n}^{(2)}}  \label{nunu21}
\end{equation}%
Next, it follows from Lemma \ref{l:redu} that%
\begin{equation}
\nu _{M_{n}^{(2)}}=\nu _{M_{n-1}^{(3)}}+O(1/n),\;n\rightarrow \infty ,
\label{nunu3}
\end{equation}%
where
\begin{equation}
M_{n-1}^{(3)}=[D_{n}^{V_{n}}(||x_{n}||V_{n}e_{n}+b_{n})]
O_{n-1}[R_{n}^{O_{n}(x)}]O_{n-1}^{T}[D_{n}^{V_{n}}(||x||V_{n}e_{n}+b_{n})].
\label{mo3}
\end{equation}
and for any $n\times n$ matrix $A$ we denote $[A]$ its $(n-1)\times (n-1)$
upper left block. Hence, $M_{n-1}^{(3)}$ is a $(n-1)\times (n-1)$ matrix.

Combining (\ref{nunu1}) -- (\ref{nunu3}), we obtain
\begin{equation*}
\nu _{M_{n}^{(1)}}=\nu _{M_{n-1}^{(3)}}+O(1/n),\;n\rightarrow \infty .
\end{equation*}%
It is important that $V_{n}$ and $O_{n-1}=[\mathcal{O}_{n}]$ are independent
(see (\ref{orep}) -- (\ref{hmfa})) and that $V_{n}$ is present only in $D_{n}^{V_{n}}$ of (\ref{mo2}).

We conclude that the initial problem to find the probability 1 the $%
n\rightarrow \infty $ limit $\nu _{M}$ of the NCM $\nu _{M_{n}}$ of the
matrix (\ref{mncal}) reduces to the traditional problem of random matrix
theory (see e.g. \cite{Pa-Sh:11}, Section 10,\cite{Pa-Va:22,Va:01} and Lemma \ref%
{l:vas} below) to find the $n\rightarrow \infty $ limit of the NCM of a
particular case $M_{n}^{(3)}$of (\ref{mnsf}) where the role of $\mathsf{K}%
_{n}=\mathsf{D}_{n}^{2}$ and $\mathsf{R}_{n}$ play $[\mathcal{R}_{n+1}]$ and
$[\mathcal{D}_{n+1}]$ respectively with $(n+1)\times (n+1)$ matrices%
\begin{equation}
\mathcal{D}_{n+1}=D_{n+1}^{V_{n+1}}(||x||V_{n+1}e_{n+1}+b_{n+1}),\;\mathcal{R%
}_{n+1}=R_{n+1}^{O_{n+1}(x)},  \label{drcal}
\end{equation}%
see (\ref{mo1}) and (\ref{mo2}).

It follows from (\ref{drcal}), Lemma \ref{l:rank}, (\ref{mo1}) and the
orthogonal invariance of the NCM of any real symmetric matrix that%
\begin{equation*}
\nu _{[\mathcal{R}_{n+1}]}=\nu _{\mathcal{R}_{n+1}}+O(1/n)=\nu
_{R_{n+1}}+O(1/n).
\end{equation*}%
Thus, we obtain the convergence of $\nu _{[\mathcal{R}_{n+1}]}$
with probability 1 of $\nu _{[\mathcal{R}_{n+1}]}$
to $\nu _{R}$ given by (\ref{nur}).

Likewise, the NCM of $[\mathcal{D}_{n+1}]$ equals the NCM\ of $\mathcal{D}%
_{n+1}$ in (\ref{drcal}) up to $O(1/n)$ and the NCM of $\mathcal{D}_{n+1}$
equals the NCM of $D_{n+1}(||x||V_{n+1}e_{n+1}+b_{n+1})$, because $\mathcal{D%
}_{n+1}$ is orthogonal equivalent to $D_{n+1}(||x||V_{n+1}e_{n+1}+b_{n+1})$
(see (\ref{drcal})) and (\ref{mo2})).  The validity of condition (\ref%
{lnkr}) for $\nu _{D_{n+1}]}$ and the explicit form of the limiting measure
follow from Lemma \ref{l:xlyl}.

As for condition (\ref{nuam}), it is valid because of (\ref{r2}) for $\nu _{\lbrack \mathcal{R}_{n+1}]}$ and because of (\ref{nuka})
and (\ref{Phi}) for $\nu _{\lbrack \mathcal{D}_{n+1}]}$.%
\end{proof}

\begin{remark}
\label{r:kgen} As is noted at the beginning of Section 2, despite the fact
that the matrices $D^{l}$ of (\ref{D}), hence $K_{n}^{l}$ of (\ref{kan}),
are random and \emph{depend} on $O^{l}$ of (\ref{wga}), the limiting
eigenvalue distribution of $M_{n}^{L}$ of (\ref{JJM}) corresponds to the
case where the analogs $\mathsf{D}_n$ of $D^{l}_n$ of (\ref{D}) are random
but \emph{independent} of $O^{l}$ as in (\ref{mnsf}), see (\ref{nukal}) and (%
\ref{nuka}). The emergence of this remarkable property of $M_{n}^{L}$ is
well seen in the above proof. 
\end{remark}

Theorem \ref{t:ind} yields an analytic form of the binary operation (\ref%
{opdia}) of the free multiplicative convolution via equations (\ref{fhk}) --
(\ref{kh}).
It is convenient to write the equations in a compact form analogous to that
of free probability theory \cite{Ch-Co:18,Mi-Sp:17}. This, in particular,
makes explicit the symmetry and the transitivity of the operation.

\begin{corollary}
\label{c:conv} Let $\nu _{A},\;A=K,R,M$ be the probability measures
(non-negative measures of the total mass 1) entering (\ref{fhk}) -- (\ref{kh}%
) and $m_{A},\;A=K,R,M$ be their moment generating functions (see (\ref{mgen}%
) -- (\ref{stmg})). Then:

(i) the functional inverses $z_{A},\;A=K,R,M$ of $m_{A},\;A=K,R,M$ are related as follows
\begin{equation}
z_{M}(m)=z_{K}(m)z_{R}(m)(1+m)m^{-1};  \label{mconv}
\end{equation}

(ii) if
\begin{equation}
S_{A}(m)=z_{A}(m)(m+1)/m,\;A=K,R,M,  \label{str}
\end{equation}%
is the S-transform of $\nu _{A}$ \cite{Mi-Sp:17}, \ then%
\begin{equation}
S_{M}(m)=S_{K}(m)S_{R}(m)  \label{sconv}
\end{equation}%
i.e., according to the terminology of free probability theory, $\nu _{M}$ is
the free multiplicative convolution of $\nu _{K}$ and $\nu _{R}$ (see (\ref%
{opdia})).
\end{corollary}


\begin{proof}
Recall that given a non-negative measure $\nu ,\;\nu (\mathbb{R})=1$, its
Stieltjes transform $f_{\nu }$ (see (\ref{stm}) and its moment generating
function $m_{\nu }$ (see (\ref{mgen}) are related as (cf. \ref{stmg}))%
\begin{equation}
zf_{\nu }(z)=-(m_{\nu }(z^{-1})+1).  \label{fmb}
\end{equation}%
Note that the moment generating function is well defined by this formula even
if $\nu $ has no finite moment of sufficiently high order. Besides, the
formula shows that the functional inverse $z_{\nu }$ of $m_{\nu }$ is well
defined in a neighborhood of the origin of the $m$-plane where $|\Im m|\geq
\varepsilon |\Re m|$ for some $\varepsilon >0$, because it follows from (\ref%
{stm}) that
\begin{equation}
f_{\nu }=z^{-1}(1+o(d^{-1})),\;f_{\nu }^{\prime }=z^{-2}(1+o(d^{-1})),\;d=%
\mathrm{dist}(z,\mathbb{R}_{+})\rightarrow \infty .  \label{fasy}
\end{equation}

Let us prove (\ref{mconv}). By using (\ref{fmb}) for $\nu =\nu _{A},\;A=M,R$
to pass from $f=f_{A},\;A=M,R$ to $m=m_{A},\;A=M,R$ in (\ref{hk}), we obtain
\begin{equation*}
m_{M}(z^{-1})=m_{R}(h_{K}(z)/(m_{M}(z^{-1})+1)).
\end{equation*}%
An analogous argument for $A=M,K$, applied to (\ref{kh}), yields%
\begin{equation*}
m_{M}(z^{-1})=m_{K}(h_{R}(z)/(m_{M}(z^{-1})+1)).
\end{equation*}%
Changing in the both relations $z\rightarrow z^{-1}$ and applying to the
first one the functional inverse $z_{R}$ of $m_{R}$ and to the second one
the functional inverse $z_{K}$ of $m_{K}$, we get%
\begin{equation*}
(1+m)z_{R}(m)=h_{K}(z_{M}^{-1}(m)),
\;(1+m)z_{K}(m)=h_{R}(z_{M}^{-1}(m)),
\end{equation*}
hence,
\[
(1+m)^2 z_{K}(m)z_{R}(m)=h_{K}(z_{M}^{-1}(m))h_{R}(z_{M}^{-1}(m)).
\]
Now we use again (\ref{fmb}) to write equation (\ref{fhk}) as%
\begin{equation*}
z_{M}(m)m(m+1)=h_{K}(z_{M}^{-1}(m))h_{R}(z_{M}^{-1}(m)).
\end{equation*}%
Combining the last two relations, we obtain (\ref{mconv}).

To obtain (\ref{sconv}) we combine (\ref{str}) and (\ref{mconv}).
\end{proof}

\bigskip

We will now give the list of results on orthogonal matrices, linear algebra
and random matrix theory that are used in the proof of the theorem.

\smallskip
First is a collection of facts on the group $SO(n)$.
\begin{proposition}
\label{p:facts} Given a positive integer $n$ consider the group $SO(n)$ of $%
n\times n$ orthogonal matrices with determinant 1. The following facts on $%
SO(n)$ are valid.

\medskip (i) Viewing $O_{n}\in SO(n)$ as the orthogonal transformation of $%
\mathbb{R}^{n}$ with an orthogonal basis $\{e_{j}\}_{j=1}^{n}$ and denoting $%
g_{k}(\theta )$ the rotation by the angle $\theta $ in the $(e_{k+1},e_{k})$
plane from $e_{k+1}$ to $e_{k}$ and%
\begin{equation*}
g^{(k)}=g_{1}(\theta _{1}^{k})\dots g_{1}(\theta _{k}^{k}),~\theta
_{1}^{k}\in \lbrack 0,2\pi ),~\theta _{j}^{k}\in \lbrack 0,\pi ),~j\neq 1,
\end{equation*}%
we have
\begin{equation}
O_{n}=V_{n}\mathcal{O}_{n},  \label{orep}
\end{equation}%
where
\begin{equation}
V_{n}=g^{(n-1)}=:V_{n}(\Theta _{1}),\;\mathcal{O}_{n}=g^{(n-2)} \dots
g^{(1)}=:\mathcal{O}_{n}(\Theta _{2}),  \label{ovo}
\end{equation}%
i.e., $V_{n}$ and $\mathcal{O}_{n}$ depend only on%
\begin{equation}
\Theta _{n}^{(1)}=\{\theta _{j}^{(n-1)}\}_{j=1}^{n-1},~\Theta
_{n}^{(2)}=\{\theta _{j}^{(k)}\}_{j,k=1}^{j=k,k=n-1}  \label{tets}
\end{equation}%
respectively, i.e., on the independent parametrization of the unit sphere $S^{n-1}$ and
of that of the group $SO(n-1)$, and $\mathcal{O}_{n}$ is the block-diagonal
matrix whose $(n-1)\times (n-1)$ upper left block is a $SO(n-1)$ matrix and
lower right $1\times 1$ block is $1$:%
\begin{equation}
\mathcal{O}_{n}=O_{n-1}\oplus \mathbf{1}_{1}=\left(
\begin{array}{cc}
O_{n-1} & 0 \\
0 & 1%
\end{array}%
\right);  \label{ocal}
\end{equation}

\medskip (ii) If $dO_{n}$ is the normalized Haar measure of $SO(n)$, then%
\begin{equation}
dO_{n}=dV_{n}dO_{n-1},  \label{hmfa}
\end{equation}%
where $dV_{n}$ is the normalized measure on the manifold determined by $%
\Theta _{1}$, in fact the "uniform" probability distribution of the vector
\begin{equation}
\xi _{n}=O_{n}e_{n}  \label{xin}
\end{equation}%
$\,$\ over $\mathbb{S}^{n-1}$ and $dO_{n-1}$ is a probability measure on $%
SO(n-1)$, in fact its normalized Haar measure of $%
SO(n-1)$;

\medskip (iii) Let $\Phi :SO(n)\rightarrow \mathcal{M}_{n}(\mathbb{C})$ be a
map admitting a $C^{1}$ continuation into an open neighborhood of $SO(n)$ in
the whole algebra $\mathcal{M}_{n}(\mathbb{R})$ of $n\times n$ matrices and $%
\mathbf{E}_{n}\{\dots \}$ denotes the integration (expectation) with respect
to the normalized to unity Haar measure on $SO(n)$. Then we have%
\begin{equation}
\mathbf{E}_{n}\{\Phi ^{\prime }(O_{n})\cdot A_{n}O_{n}\}=0,\;\mathbf{E}%
_{n}\{\Phi ^{\prime }(O_{n})\cdot O_{n}A_{n}\}=0,\quad \forall A_{n}\in
\mathcal{A}_{n},  \label{diffO}
\end{equation}%
where $\mathcal{A}_{n}$ is the space of $n\times n$ real antisymmetric
matrices and $\Phi ^{\prime }$ is viewed as a linear map from $\mathcal{M}%
_{n}(\mathbb{R})$ to $\mathcal{M}_{n}(\mathbb{C)}$, or, in the coordinate
form
\begin{equation}
\sum_{l=1}^{n}\mathbf{E}_n\{\Phi _{lj}^{\prime }(O_{n})(O_{n})_{lk}-\Phi
_{lk}^{\prime }(O_{n})(O_{n})_{lj}\}=0,\quad j,k=1,\dots ,n,  \label{diffOc}
\end{equation}%
%
where
\begin{align}
\Phi _{jk}^{\prime }(U) =\Phi ^{\prime }(U)\cdot \mathcal{E}%
^{(jk)}=\lim_{\varepsilon \rightarrow \infty }(\Phi (U+\varepsilon \mathcal{E%
}^{(jk)}))-\Phi (U))\varepsilon ^{-1},  \notag \\
\mathcal{E}^{(jk)} =\{\mathcal{E}_{ab}^{(jk)}\}_{a,b=1}^{n},\;\mathcal{E}%
_{ab}^{(jk)}=\delta _{aj}\delta _{bk}\in \mathcal{M}_{n}(\mathbb{R}),
\label{derph}
\end{align}%
i.e., $\{\mathcal{E}^{(jk)}\}_{j,k=1}^{n}$ is a basis in $\mathcal{M}_{n}(%
\mathbb{R})$;

\medskip (iv) We have in the above notation for a map $\varphi :
SO(n)\rightarrow \mathbb{C}$ and a sufficiently large $n$
\begin{align}
&\mathbf{Var}_{n}\{\varphi \} :=\mathbf{E}_{n}\{|\varphi |^{2}\}-|\mathbf{E}%
_{n}\{\varphi \}|^{2}  \notag \\
&\hspace{1.5cm}\leq \frac{C}{n}\mathbf{E}_{n}\Big\{\sum_{1\leq j<k\leq
n}^{n}\left\vert \varphi _{jk}^{\prime }\right\vert ^{2}\Big\},  \label{PNOF}
\end{align}%
where $C$ is an absolute constant and
\begin{align}
&\varphi _{jk}^{\prime }(O_{n}) =\varphi ^{\prime }(O_{n})\cdot
A^{(jk)}=\lim_{\varepsilon \rightarrow \infty }(\varphi (O_{n}(\mathbf{1}%
_{n}+\varepsilon A^{(jk)}))-\varphi (O_{n}))\varepsilon ^{-1},  \notag \\
&A^{(jk)} =\{A_{ab}^{(jk)}\}_{a,b=1}^{n},\;A_{ab}^{(jk)}=\delta _{aj}\delta
_{bk}-\delta _{ak}\delta _{bj},  \label{deras}
\end{align}%
i.e., $\{A^{(jk)}\}_{0\leq j<k\leq n}$ is the basis of $\mathcal{A}_{n}$;

\medskip (v) There exists an infinite-dimensional probability space $\Omega
_{O}$ on which all $O_{n},~n\geq 1$, are simultaneously defined.
\end{proposition}

Items (i) and (ii) of proposition are structure properties of $SO(n)$, see
\cite{Vi:68}, Section IX.1 and \cite{Mu:62}, Chapters 2 and 5. Item (iii)
follows from the invariance of the Haar measure of $SO(n)$ with respect to
the left $O\rightarrow e^{\varepsilon A}O$ and the right $O\rightarrow
Oe^{\varepsilon A}$ shifts with $\varepsilon \rightarrow 0$, see \cite%
{Pa-Sh:11}, Section 8.1. Item (iv) is a version of the Poincar\'{e}
inequality for $SO(n)$, see \cite{Pa-Sh:11}, Section 8.1 and item (v) is
again a structure property of $SO(n)$, see \cite{Ne:11}, Section 2.10 and
\cite{Pa-Sh:11}, Section 8.1. 

\begin{lemma}
\label{l:rank} Let $T_{n}$ be an $n\times n$ matrix and $[T_{n}]$ be its
upper left block, i.e., if $P_{n}$ is the orthogonal projection on the last
basis vector $e_{n}$ of $\ \mathbb{C}^{n}$ and $Q_{n}$ is the complementary
orthogonal projection, so that%
\begin{equation}
Q_{n}+P_{n}=\mathbf{1}_{n},  \label{qp}
\end{equation}%
we can write%
\begin{equation}\label{tq}
[T_{n}]=[T_{Q_{n}}],\;T_{Q_{n}}:=Q_{n}T_{n}Q_{n}=\left(
\begin{array}{cc}
\lbrack T_{n}] & 0 \\
0 & 0%
\end{array}%
\right) .
\end{equation}%
We have:

(i) if $A_{n}$ is an $n\times n$ hermitian matrix, then%
\begin{equation}
A_{n}=A_{Q_n}+R_{3},\;\mathrm{rank\,}R_{3}\leq 3;  \label{ar3}
\end{equation}

(ii) if $A_{n}$ and $B_{n}$ are $n\times n$ hermitian matrices, $X_{n}$ is
an $n\times n$ block-diagonal matrix%
\begin{equation}
X_{n}=\left(
\begin{array}{cc}
\lbrack X_{n}] & 0 \\
0 & 1%
\end{array}%
\right) =X_{Q_{n}}+P_{n},~X_{Q_{n}}=Q_{n}X_{n}Q_{n}  \label{xpq}
\end{equation}%
and%
\begin{equation}
C_{n}=A_{n}X_{n}^{\ast }B_{n}X_{n}A_{n},  \label{c}
\end{equation}%
then%
\begin{equation}
C_{n}=\mathcal{C}_{Q_{n}}+R_{6},  \label{cqr6}
\end{equation}%
where%
\begin{equation}
\mathcal{C}_{Q_{n}}=\left(
\begin{array}{cc}
\lbrack \mathcal{C}_{Q_{n}}] & 0 \\
0 & 0%
\end{array}%
\right) ,\;[\mathcal{C}_{Q_{n}}]=[A_{n}][X_{n}^{\ast }][B_{n}][X_{n}][A_{n}]~
\label{cq}
\end{equation}%
and%
\begin{equation}
\mathrm{rank\,}R_{6}\leq 6.  \label{er6}
\end{equation}
\end{lemma}

\begin{proof}
(i) We will omit the subindex $n$ in the cases where it does not lead to
confusion. We have from (\ref{qp}) and (\ref{tq})
\begin{equation}
A=A_{Q}+QAP+PAQ+PAP.  \label{aex}
\end{equation}%
Given $x,y\in \mathbb{C}^{n}$, define the rank-one matrix
\begin{equation}
L_{xy}u:=(u,x)y,~\forall u\in \mathbb{C}^{n}.  \label{lxy}
\end{equation}%
We obtain
\begin{eqnarray*}
QAP &=&L_{e_{n},QAe_{n}},~PAQ=L_{QAe_{n},e_{n}},~ \\
PAP &=&(Ae_{n},e_{n})L_{e_{n},e_{n}}=(Ae_{n},e_{n})P,
\end{eqnarray*}%
implying that $A-A_{Q}$ is of rank 3 at most. This and the min-max principle
of linear algebra yields (\ref{ar3}).

(ii) Applying (\ref{ar3}) to $C$ of (\ref{c}), we get%
\begin{equation}
C=C_{Q}+R_{3},\;\mathrm{rank}\,R_{3}\leq 3.  \label{cr3}
\end{equation}%
Next, we have from (\ref{qp}), (\ref{xpq}), (\ref{c}) and (\ref{cq}) \
\begin{equation}
C_{Q}=\mathcal{C}_{Q}+T_{1}+T_{2}+T_{3},  \label{ct}
\end{equation}%
where,
\begin{equation}
\mathcal{C}_{Q}=A_{Q}X_{Q}^{\ast }B_{Q}X_{Q}A_{Q}  \label{cq0}
\end{equation}%
and, in view of (\ref{lxy}),%
\begin{align}
& T_{1}:=A_{Q}X_{Q}^{\ast }BPAQ=L_{QAe_{n},A_{Q}X_{Q}^{\ast }Be_{n}},  \notag
\\
& T_{2}:=QAPBX_{Q}A_{Q}=L_{A_{Q}X_{Q}^{\ast }Be_{n},QAe_{n},}  \notag \\
& T_{3}:=QAPBPAQ=(Be_{n},e_{n})L_{QAe_{n},QAe_{n}}.  \label{t123}
\end{align}%
All the factors on the right of (\ref{cq0}) are of the form $T_{Q}$ in (\ref{tq}), $\;T=A,B,X$, hence the l.h.s. $\mathcal{C}_{Q}$ is also of this form
with $[\mathcal{C}_{Q}]$ given by the r.h.s. of (\ref{cq}). In addition,
since each $T_{a},\;a=1,2,3$ in (\ref{t123}) is of rank 1, we have that $%
\mathrm{rank}(T_{1}+T_{2}+T_{3})\leq 3$. This, the min-max principle, (\ref%
{cr3}) and (\ref{ct}) imply (\ref{er6}).
\end{proof}

\begin{lemma}
\label{l:redu} Let $M_{n}$ be defined in (\ref{mncal}) -- (\ref{qqn}) and
let
\begin{equation}
\mathbf{M}_{n-1}=[S_{n}]O_{n-1}^{T}[\widehat{K}_{n}]O_{n-1}[S_{n}]
\label{mnbf}
\end{equation}%
be $(n-1)\times (n-1)$ matrix, where $[S_{n}]$ is the $(n-1)\times (n-1)$
upper left block of $S_{n}$, $O_{n-1}$ is the Haar distributed $SO(n-1)$
matrix, $[K_{n}^{V_{n}}]$ is the $(n-1)\times (n-1)$ upper left block of $\
K_{n}^{V_{n}}:=(D_{n}^{V_{n}})^{2}=V_{n}^{T}K_{n}V_{n}$, where $V_{n}$ is
given (\ref{orep}) -- (\ref{tets}) and $K_{n}$ is given by (\ref{Dn}).
Denote $\nu _{M_{n}}$ and $\nu _{\mathbf{M}_{n-1}}$ the NCM of $M_{n}$ and $%
\mathbf{M}_{n-1}$. Then 
\begin{equation}
\nu _{M_{n}}(\Delta )-\nu _{\mathbf{M}_{n}}(\Delta )=O(1/n),\;n\rightarrow
\infty .  \label{nn7}
\end{equation}
\end{lemma}

\begin{proof}
According to Proposition \ref{p:facts}, we have $O_{n}=V_{n}\mathcal{O}_{n}$%
, where $V_{n}$ and $\mathcal{O}_{n}$ are $n\times n$ random orthogonal
matrix given by (\ref{ovo}), (\ref{tets}) and (\ref{ocal}).

This and (\ref{mncal}) allow us to write%
\begin{equation}
M_{n}=S_{n}\mathcal{O}_{n}^{T}K_{n}^{V_{n}}\mathcal{O}_{n}S_{n},%
\;K_{n}^{V_{n}}=V_{n}^{T}K_{n}V_{n}.  \label{mnbof}
\end{equation}%
Next, we use Lemma \ref{l:rank} with $A=S_{n},\;B_{n}=K_{n}^{V_{n}}$ and $%
X_{n}=\mathcal{O}_{n}$, hence, $C_{n}=M_{n}$, implying in the notation of
the lemma%
\begin{equation}
M_{n}=(M_{n})_{Q_{n}}+R_{6},\;\mathrm{rank\,}R_{6}\leq 6.  \label{mn6}
\end{equation}%
Thus, if $\nu _{M_{n}}$ and $\nu _{(M_{n})_{Q}}$ are the NCM of $M_{n}$ and $%
(M_{n})_{Q}$, then it follows from (\ref{mn6}) and the min-max principle of
linear algebra that%
\begin{equation}
\nu _{M_{n}}-\nu _{(M_{n})_{Q_{n}}}=O(1/n),\;n\rightarrow \infty .
\label{num6}
\end{equation}%
Next, according to the same lemma, we have for the $(n-1)\times (n-1)$ upper
left block $[(M_{n})_{Q}]$ of $(M_{n})_{Q}$
\begin{equation}
\lbrack (M_{n})_{Q}]=[S_{n}]O_{n-1}^{T}[K_{n}^{V_{n}}]O_{n-1}[S_{n}].
\label{mnpb}
\end{equation}%
Thus, if $\nu _{M_{n}}$ and $\nu _{\lbrack (M_{n})_{Q}]}$ are the NCM of the
$n\times n$ matrix$\ (M_{n})_{Q}$ and $(n-1)\times (n-1)$ matrix $%
[(M_{n})_{Q}]$, see (\ref{cq}), then%
\begin{equation*}
\nu _{(M_{n})_{Q}}=\frac{n-1}{n}\nu _{\lbrack (M_{n})_{Q}]}=\nu _{\lbrack
(M_{n})_{Q}]}-n^{-1}\nu _{\lbrack (M_{n})_{Q}]},
\end{equation*}%
and since $0\leq \nu _{\lbrack (M_{n})_{Q}]}(\Delta )\leq 1$ uniformly in $%
\Delta \subset \mathbb{R}_{+}$, we have%
\begin{equation}
\nu _{(M_{n})_{Q}}-\nu _{\lbrack (M_{n})_{Q}]}=O(1/n),\;n\rightarrow \infty .
\label{num1}
\end{equation}%
Combining (\ref{num6}) and (\ref{num1}), we get%
\begin{equation*}
|\nu _{M_{n}}(\Delta )-\nu _{\lbrack (M_{n})_{Q}]}=O(1/n),\;n\rightarrow
\infty .
\end{equation*}%
Denoting now $[(M_{n})_{Q}]:=\mathbf{M}_{n-1}$ and using (\ref{mnpb}), we
obtain (\ref{mnbf}).
\end{proof}

\smallskip The next lemma gives am explicit analytic form of the operation (%
\ref{opdia}).

\begin{lemma}
\label{l:vas} Let $\mathsf{M}_{n}$ be the $n\times n$ random matrix (\ref%
{mnsf}) where $O_{n}$ is Haar distributed over $SO(n)$ and $\mathsf{K}_{n}$
and $\mathsf{R}_{n}$ are $n\times n$ positive definite random matrices independent of $O_{n}$ and such that their Normalized Counting Measures $%
\nu _{\mathsf{K}_{n}}$ and $\nu _{\mathsf{R}_{n}}$ converge weakly with
probability 1 as $n\rightarrow \infty $ to the non-random limits $\nu _{%
\mathsf{K}}$ and $\nu _{\mathsf{R}}$ of (\ref{lnkr}) with $\nu _{\mathsf{K}}(%
\mathbb{R}_{+})=\nu _{\mathsf{R}}(\mathbb{R}_{+})=1$ and
\begin{equation}
\sup_{n} \int_{0}^{\infty }\lambda ^{2}\nu _{\mathsf{A}_n}(d\lambda )<\infty ,\;%
\mathsf{A}=\mathsf{K,R}.  \label{nuam}
\end{equation}%
Then the Normalized Counting Measure $\nu _{\mathsf{M}_{n}}$ of  $\mathsf{M}%
_{n}$ converges weakly with probability 1 as $n\rightarrow \infty $ to a
non-random limit $\nu _{\mathsf{M}}$ and its
Stieltjes transform (see (\ref{fm})) is a unique solution of  (\ref{fhk}) -- (\ref{kh}) satisfying (%
\ref{hcond}).
\end{lemma}

The lemma is known in fact, see \cite{Mi-Sp:17} Section 4.3, \cite{Pa-Sh:11}, Section 10.4 and \cite{Va:01} for unitary matrices.
A streamlined proof applicable to both unitary and orthogonal matrices
is given in  \cite{Pa-Va:22}.


\medskip The next two lemmas deal with asymptotic properties of the
activations vectors  $x^{l}$ in the $l$th layer, see (\ref{rec}). It is an
extended version (treating the convergence with probability 1) of assertions
proved in \cite{Ma-Co:16,Po-Co:16,Sc-Co:17} for expectations.

\medskip
The first lemma is a version of the Law of Large Numbers for random vectors
that are uniformly distributed over $\mathbb{S}^{n-1}$.

\begin{lemma}
\label{l:clt} Let $\chi :\mathbb{R}\rightarrow \mathbb{R}$ be piece-wise
continuous and%
\begin{equation}
\sup_{x\in \mathbb{R}}|\chi (x)|=\widehat\chi  <\infty ,  \label{chi01}
\end{equation}%
$\xi _{n}=\{\xi _{jn}\}_{j=1}^{n}\in \mathbb{S}^{n-1}$ be the random vector
uniformly distributed over the sphere $\mathbb{S}^{n-1}$, \ $%
b=\{b_{j}\}_{j=1}^{\infty }$ (cf. (\ref%
{binf})) be a collection of i.i.d. Gaussian random
variables of zero mean and variance $\sigma _{b}^{2}$ and $%
\{a_{n}\}_{n=1}^{\infty }$ be a real valued sequence such that
\begin{equation}
\lim_{n\rightarrow \infty }a_{n}=a.  \label{ana}
\end{equation}%
%
%
%
Denoting  $\Omega _{b}$  the probability space generated by $b$ and writing $\xi _{n}=O_{n}e_{n}$, we can say that $b$ and $\{\xi
_{n}\}_{n=1}^{\infty }$ are defined on the same probability space $\Omega
_{Ob}=\Omega _{O}\times \Omega _{b}$ (cf. (\ref{binf}) -- (\ref{xinf})). Set
\begin{equation}
\chi _{n}=\frac{1}{n}\sum_{j=1}^{n}\chi \left( n^{1/2}a_{n}\xi
_{jn}+b_{j}\right) .  \label{chin}
\end{equation}%
and view it as a random variable in $\Omega _{Ob}$. Then we have with
probability 1 in $\Omega _{Ob}$
\begin{equation}
\chi =\lim_{n\rightarrow \infty }\chi _{n}=\int_{-\infty }^{\infty }\chi
\left( a\gamma +b\right) \frac{e^{-\gamma ^{2}/2-b^{2}/2\sigma _{b}^{2}}}{%
2\pi \sigma _{b}}d\gamma db.  \label{chil}
\end{equation}
\end{lemma}


\begin{proof}
Denote $\mathbf{E}_{Ob}\{\ldots \}$ the expectation in
$\Omega _{Ob}$ and $\mathbf{E}_{b}\{\ldots \}$ \ and $\mathbf{E}_{O}\{\ldots
\}$ the expectation in $\Omega _{b}$ and $\Omega _{O}$ respectively, so that
$\mathbf{E}_{Ob}=\mathbf{E}_{O}\mathbf{E}_{b}$. Write%
\begin{equation}
\chi _{n}=T_{1n}+T_{2n},  \label{chitt}
\end{equation}%
where%
\begin{eqnarray}
T_{1n} &=&\chi _{n}-\mathbf{E}_{b}\{\chi _{n}\}  \notag \\
&=&\frac{1}{n}\sum_{j=1}^{n}\chi (n^{1/2}a_{n}\xi _{jn}+b_{j})-\frac{1}{n}%
\sum_{j=1}^{n}\mathbf{E}_b\{\chi (n^{1/2}a_{n}\xi _{jn}+b_{j}\}  \label{t1}
\end{eqnarray}%
and%
\begin{equation}
T_{2n}=\mathbf{E}_{b}\{\chi _{n}\}=\frac{1}{n}\sum_{j=1}^{n}\mathbf{E}%
_{b}\{\chi \left( n^{1/2}a_{n}\xi _{jn}+b_{j}\right) \}.  \label{t2}
\end{equation}%
For any fixed $\{\xi _{jn}\}_{j=1}^{n}\in \mathbb{S}^{n-1}$
the r.h.s. of (%
\ref{t1}) is the arithmetic mean of the bounded (see (\ref{chi01}) i.i.d.
random variables of zero mean in $\Omega _{b}$. It follows then from a
standard calculation and (\ref{chi01}) that 
\begin{equation*}
\mathbf{E}_{b}\{|T_{1n}|^{4}\}\leq 3 \widehat\chi _{0}^{4}/n^{2}.
\end{equation*}%
Since the r.h.s. of this bound is independent of $\{\xi _{jn}\}_{j=1}^{n}\in
\mathbb{S}^{n-1}$, we obtain
\begin{equation*}
\mathbf{E}_{Ob}\{|T_{1n}|^{4}\}\leq 3\widehat{\chi}
_{0}^{4}/n^{2}.
\end{equation*}%
This and the Borel-Cantelli lemma imply that the limit%
\begin{equation}
\lim_{n\rightarrow \infty }T_{1n}=0.  \label{lt1}
\end{equation}%
holds with probability 1 in $\mathbf{E}_{Ob}$.

Let us prove now that the almost sure limit of $T_{2n}$ of (\ref{t2}) equals
the r.h.s. of (\ref{chil}). To this end we first rewrite $T_{2n}$ as%
\begin{equation}
T_{2n}=\frac{1}{n}\sum_{j=1}^{n}\overline{\chi }\left( n^{1/2}a\xi
_{jn}\right) +T_{3n}  \label{t22}
\end{equation}%
where%
\begin{equation}
\overline{\chi }(x)=\mathbf{E}_{b}\{\chi (x+b_{j})\}=\int_{-\infty }^{\infty
}\chi \left( x+b\right) \frac{e^{-b^{2}/\sigma _{b}^{2}}}{\sqrt{2\pi \sigma
_{b}^{2}}}db  \label{chibx}
\end{equation}%
and%
\begin{equation}
T_{3n}=\frac{1}{n}\sum_{j=1}^{n}(\overline{\chi }\left( n^{1/2}a_{n}\xi
_{jn}\right) -\overline{\chi }\left( n^{1/2}a\xi _{jn}\right) ).  \label{t3}
\end{equation}%
It follows from (\ref{chi01}) and (\ref{chibx}) that
\begin{equation}
\sup_{x\in \mathbb{R}}|\overline{\chi }^{\prime }(x)|:=\chi _{1}<\infty
,\;\chi _{1}=(2/\pi \sigma _{b}^{2})^{1/2}\widehat\chi _{0},  \label{chi11}
\end{equation}%
hence, we have for $\{\xi _{jn}\}_{j=1}^{n}\in \mathbb{S}^{n-1}$ by Schwarz
inequality
\begin{align}
& |T_{3n}|\leq \chi _{1}n^{1/2}|a_{n}-a| \, \frac{1}{n}
\sum_{j=1}^{n}|\xi _{jn}|
\notag \\
& \hspace{0.5cm}\leq \chi _{1}n^{1/2}|a_{n}-a|
\left( \frac{1}{n} \sum_{j=1}^{n}|\xi _{jn}|^{2}\right) ^{2}=\chi _{1}|a_{n}-a|.
\label{t3n}
\end{align}%
Next, it follows from a direct calculation that if $\{\gamma _{j}\}_{j=1}^{n}
$ is the collection of independent standard Gaussian random variables and%
\begin{equation}
\Gamma _{n}^{2}=\sum_{j=1}^{n}\gamma _{j}^{2},  \label{Gan}
\end{equation}%
then%
\begin{equation}
\xi _{jn}=\gamma _{j}/\Gamma _{n},\;j=1,\ldots n  \label{xigan}
\end{equation}%
(we thank A. Sodin for the indication on this nice fact).

Thus, we can write in view of (\ref{ana}), (\ref{t22}), (\ref{t3n}) and (\ref%
{xigan})%
\begin{eqnarray}
T_{2n} &=&n^{-1}\sum_{j=1}^{n}\overline{\chi }\left( an^{1/2}\gamma
_{j}/\Gamma _{n}\right) +o(1)  \notag \\
&=&n^{-1}\sum_{j=1}^{n}\overline{\chi }\left( a\gamma _{j}\right)
+T_{4n}+o(1),\;n\rightarrow \infty   \label{t23}
\end{eqnarray}%
where%
\begin{equation*}
T_{4n}=n^{-1}\sum_{j=1}^{n}\left( \overline{\chi }\left( an^{1/2}\gamma
_{j}/\Gamma _{n}\right) -\overline{\chi }\left( a\gamma _{j}\right) \right) .
\end{equation*}%
Repeating the argument leading to (\ref{t3n}), we obtain
\begin{eqnarray*}
|T_{4n}| &\leq &\chi _{1}a|n^{1/2}/\Gamma _{n}-1|n^{-1}\sum_{j=1}^{n}|\gamma
_{j}| \\
&\leq &\chi _{1}|1-\Gamma _{n}/n^{1/2}|
\end{eqnarray*}%
According to the strong Law of Large Numbers and (\ref{Gan}), we have with
probability 1
\begin{equation}
\lim_{n\rightarrow \infty }\Gamma _{n}/n^{1/2}=\lim_{n\rightarrow \infty
}\left( n^{-1}\sum_{j=1}^{n}|\gamma _{j}|^{2}\right) ^{1/2}=(\mathbf{E}%
\{\gamma _{1}^{2}\})^{1/2}=1,  \label{lgan}
\end{equation}%
hence
\begin{equation*}
\lim_{n\rightarrow \infty }T_{4n}=0
\end{equation*}%
with probability 1.

We are left with the proof that the first term of (\ref{t23}) tends to the
r.h.s. of (\ref{chil}) with probability 1 as $n\rightarrow \infty $. This
follows immediately from the strong Law of Large Numbers since $\{\gamma
_{j}\}_{j=1}^{\infty }$ are the independent standard Gaussian random
variables.
\end{proof}

\medskip
We will use the lemma to prove formulas (\ref{nukal}) and (\ref{qlql}).
\begin{lemma}
\label{l:xlyl} Let $y^{l}=\{y_{j}^{l}\}_{j=1}^{n},\;l=1,\dots, L$ be
post-affine random vectors defined in (\ref{rec}) -- (\ref{xl}) with $x^{0}$
satisfying (\ref{q0}), $\chi :\mathbb{R}\rightarrow \mathbb{R}$ be a bounded
piece-wise continuous function and $\Omega _{L}$ be defined in (\ref{oml}).
Set
\begin{equation}
\chi _{n}^{l}=n^{-1}\sum_{j_{l}=1}^{n}
\chi (y_{j_{l}}^{l}),\;l= 1, \dots, L.
\label{chiln}
\end{equation}%
Then there exists $\overline{\Omega }_{l}\subset \Omega _{L},\;\mathbf{P}(%
\overline{\Omega }_{l})=1$ such that for every $\omega _{l}\in \overline{%
\Omega }_{l}$, i.e., with probability 1 in $\Omega _{L}$, the limits%
\begin{equation}
\chi ^{l}:=\lim_{n\rightarrow \infty }\chi _{n}^{l},\;l=1,\dots,L
\label{lql}
\end{equation}%
exist, are not random and equal
\begin{equation}
\chi ^{l}=\int_{-\infty }^{\infty }\chi ((q^{l})^{1/2}\gamma +b)\frac{%
e^{-\gamma ^{2}/2-b^{2}/2\sigma _{b}^{2}}}{2\pi \sigma _{b}}d\gamma
db,\;l=1,\dots L,  \label{lqg}
\end{equation}%
valid on $\overline{\Omega }_{l}$ with $q^{l}$ defined recursively as
\begin{equation}
q^{l}=\int_{-\infty }^{\infty }\varphi ^{2}((q^{l-1})^{1/2}\gamma +b)\frac{%
e^{-\gamma ^{2}/2-b^{2}/2\sigma _{b}^{2}}}{2\pi \sigma _{b}}+\sigma
_{b}^{2},\;l=1,\dots   ,L\label{ql}
\end{equation}%
with $q^{0}$ given in (\ref{q0}).

In particular, we have from the above with $\chi=\psi$ the validity with probability 1 formula (\ref{nukal}) (hence, (\ref%
{nuka}))  for the weak limit $\nu _{K^{l}}$ of the Normalized Counting
Measure $\nu _{K_{n}^{l}}$ of diagonal random matrix $K_{n}^{l}$ of (\ref%
{kan}).
\end{lemma}


\begin{proof}
Consider the case where $l=1$ in (\ref{chiln}):%
\begin{equation}
\chi _{n}^{1}=\frac{1}{n}\sum_{j_{1}=1}^{n}\chi
((O^{1}x^{0})_{j_{1}}+b_{j_{1}}^{1}).  \label{chi00}
\end{equation}%
Since probability distribution of $O^{1}$, i.e., the normalized to unity
Haar measure on $SO(n)$, is orthogonal invariant, we can replace $O^{1}$ in (%
\ref{chi00}) by $O^{1}O^{0}$ such that $O^{0}x^{0}=e_{n}$ (cf. (\ref{oxe})). Hence, the random
variable, obtained by using this replacement in (\ref{chi00}), is
stochastically equivalent to that of (\ref{chi00}), i.e., its probability
distribution coincides with that of (\ref{chi00}). We denote this random
variable again $\chi
_{n}^{1}$ and write
\begin{equation}
\chi _{n}^{1}=\frac{1}{n}\sum_{j_{0}=1}^{n}\chi (||x^{0}||\xi
_{nj_{1}}+b_{j_{1}}^{1}),  \label{chi30}
\end{equation}%
where $\{\xi _{jn}\}_{j=1}^{n}=\xi _{n}$ is the unit vector uniformly
distributed over the unit sphere $S^{n-1}$. It follows then from the
condition of the lemma that the random variable (\ref{chi30}) satisfies the
condition of Lemma \ref{l:clt} (condition (\ref{ana}) of the lemma is
guarantied by (\ref{q0})). We conclude that (\ref{lql}) -- (\ref{lqg}) hold
for $l=1$.

In particular, choosing $\chi (x)=(\varphi (x))^{2},\;\sup_{x\in \mathbb{R}%
}(\varphi (x))^{2}=\Phi _{0}^{2}$ (see (\ref{Phi})), we obtain (\ref{ql})
for $l=1$ probability 1 in $\Omega _{1}$.

Consider now the case where $l=2$. Since $\{O^{1},b^{1}\}$ and $\{O^{2},b^{2}\}$
are independent, we can fix $\omega _{1}\in \overline{\Omega }_{1},\ \mathbf{%
P}(\overline{\Omega }_{1})=1$ (a realization of $\{O^{1},b^{1}\}$) and apply
to $\chi _{n}^{2}$ of (\ref{chiln}) the same argument as that for the case $%
l=1$ above and obtain (\ref{lql}) for $l=2$ and this $\omega_1$.
This imply that to for every $\omega _{1}\in
\overline{\Omega }_{1}$ there exists $\overline{\Omega }^{2}(\omega
^{1})\subset \Omega ^{2},\;\mathbf{P}(\overline{\Omega }^{2}(\omega
^{1}))=1\,$\ on which (\ref{lql}) -- (\ref{lqg}) for $l=2$ hold. Then, by
using again the Fubini theorem, we obtain the validity of (\ref{lql}) -- (%
\ref{lqg}) for $l=2$ on a certain $\overline{\Omega }_{2}\subset \Omega
_{2}=\Omega ^{1}\otimes \Omega ^{2},\;$ $\mathbf{P}(\overline{\Omega }%
_{2})=1 $. In particular, we have (\ref{ql}) for $l=2.$ Analogous argument
applies for $l=3,4,\dots.$

To prove (\ref{nukal}) it suffices to prove the validity with probability 1
of the relation
\begin{equation*}
\lim_{n\rightarrow \infty }\int_{-\infty }^{\infty }\psi (\lambda )\nu
_{K_{n}^{l}}(d\lambda )=\int_{-\infty }^{\infty }\psi (\lambda )\nu
_{K^{l}}(d\lambda )
\end{equation*}%
for any bounded and continuous $\psi :\mathbb{R\rightarrow R}$.

In view of (\ref{rec}), (\ref{D}) and (\ref{kan}) the relation can be
written in the form%
\begin{equation*}
\lim_{n\rightarrow \infty }n^{-1}\sum_{j_{l}=1}^{n}\psi ((\varphi ^{\prime
}(y_{j_{l}}^{l}))^{2})=\int_{-\infty }^{\infty }\psi ((\varphi ^{^{\prime
}}(\gamma \sqrt{q_{l-1}}+b))^{2})\frac{e^{-\gamma ^{2}/2-b^{2}/2\sigma
_{b}^{2}}}{2\pi \sigma _{b}}d\gamma db,\;l\geq 1.
\end{equation*}%
The l.h.s. here is a particular case of (\ref{chiln}) -- (\ref{lql}) for $%
\chi =\psi \circ \varphi ^{\prime 2}$, thus, it equals the r.h.s. of (\ref%
{lqg}) for this $\chi $.
\end{proof}


\begin{remark}
\label{r:clt} Lemma \ref{l:clt} remains valid for any i.i.d. $%
\{b_{j}\}_{j=1}^{\infty }$ such that the derivative $p^{\prime }$ of the
density $p$ of their common probability law $P$ belongs to $L^{1}(\mathbb{R}%
) $. Indeed, it is easy to see that in this case we have instead of (\ref%
{chil})
\begin{equation*}
\sup_{x\in \mathbb{R}}|\overline{\chi }^{\prime }(x)|=\chi _{1}<\infty
,\;\chi _{1}=\widehat\chi _{0}||p^{\prime }||_{L^{1}(\mathbb{R})}
\end{equation*}%
and instead of (\ref{chil})%
\begin{equation*}
\chi :=\lim_{n\rightarrow \infty }\chi _{n}=\int_{-\infty }^{\infty }\chi
\left( a\gamma +b\right) \frac{e^{-\gamma ^{2}/2}}{\sqrt{2\pi }}d\gamma
P(db),
\end{equation*}%
where $P$ is the common probability law of $\{b_{j}\}_{j=1}^{\infty }$.

This leads to more general version of (\ref{lqg})
\begin{equation*}
\chi ^{l}=\int_{-\infty }^{\infty }\chi (\gamma \sqrt{q^{l-1}-\sigma _{b}^{2}%
}+b)\frac{e^{-\gamma ^{2}/2}}{\sqrt{2\pi }}P(db),\;l=1,2,\dots ,
\end{equation*}%
hence, of (\ref{ql})
\begin{equation*}
q^{l}=\int_{-\infty }^{\infty }\varphi ^{2}(\gamma \sqrt{q^{l-1}-\sigma
_{b}^{2}}+b)\frac{e^{-\gamma ^{2}/2}}{(2\pi )^{1/2}}P(db)+\sigma
_{b}^{2},\;l=2,3,\dots
\end{equation*}%
and of (\ref{nukal})%
\begin{equation*}
\nu _{K^{l}}(\Delta )=\mathbf{P}\{(\varphi ^{\prime }((q^{l-1}-\sigma
_{b}^{2})^{1/2}\gamma +b_{1}))^{2}\in \Delta \},\;\Delta \in \mathbb{R}%
,,\;l=1,\dots ,L.
\end{equation*}%
with $q^{0}$ is given by (\ref{q0}).

To have the validity of the lemma for general i.i.d. $\{b_{j}\}_{j=1}^{%
\infty }$ of zero mean and variance $\sigma _{b}^{2}$ we have to assume, for
instance, that $\chi $ has a bounded derivative.
\end{remark}


\end{document}